\documentclass[a4paper,fleqn]{cas-dc}

\usepackage[authoryear]{natbib}

\usepackage{amsmath,amssymb,amsfonts}
\usepackage{algorithm,algorithmic}
\usepackage{graphicx}
\usepackage{booktabs}
\usepackage{multirow}
\usepackage{color}
\usepackage{tabularx}
\usepackage{subcaption}
\usepackage{pifont}
\newcommand{\cmark}{\ding{51}}
\newcommand{\xmark}{\ding{55}}
\usepackage{hyperref}
\hypersetup{hidelinks=true}

\expandafter\def\csname __first_footerline:\endcsname{} 
\begin{document}
\let\WriteBookmarks\relax
\def\floatpagepagefraction{1}
\def\textpagefraction{.001}

\shorttitle{Interactive 3D brain lesion segmentation with online adaptation}
\shortauthors{Xu et~al}

\title[mode=title]{BrainIAC: Interactive 3D Brain Lesion Segmentation across Heterogeneous MRI Modalities with Online Adaptation}

\author[1]{Wentian Xu}
\fnmark[1]
\credit{Conceptualization, Methodology, Software, Investigation, Data Curation, Writing - Original Draft, Visualization} 
\author[1]{Anthony P Addison}
\fnmark[2]
\credit{Conceptualization, Methodology, Software, Validation, Investigation, Writing - Original Draft, Visualization}

\author[1]{Ziyun Liang}

\credit{Conceptualization, Validation, Writing, Visualization}

\author[1]{Harry Anthony}

\credit{Conceptualization, Validation, Writing, Visualization}

\author[1]{Guang Yang}

\credit{Conceptualization, Validation, Writing}

\author[1]{Konstantinos Kamnitsas}[orcid=https://orcid.org/0000-0003-3281-6509]
\cormark[1]

\credit{Conceptualization, Supervision, Writing - Review \& Editing}

\affiliation[1]{organization={Department of Engineering Science, University of Oxford},
                                country={UK}}

\cortext[1]{Corresponding author}
\fntext[1]{Lead author}
\fntext[2]{Key contributor}

\begin{abstract}
Brain lesion segmentation is a fundamental task in medical image analysis, playing a critical role in diagnosis, treatment planning, and longitudinal disease monitoring. Yet existing models still struggle to meet the demands of real clinical use, where deployments contain data distribution shifts, arising from differences in scanner hardware, imaging protocol (varying MRI modality sets), and new pathologies. To build a more robust 3D brain lesion segmentation model, we target three capabilities: \emph{interactive segmentation}, allowing clinicians to correct unreliable predictions through prompts such as bounding boxes and clicks; \emph{online adaptation}, enabling the model to continuously learn from these interactions and adapt to new data distributions and even learn to segment previously unseen types of brain lesion; and native support for \emph{multiple MRI modalities}. We present BrainIAC
(\underline{\textbf{Brain}} lesion
\underline{\textbf{I}}nteractive
\underline{\textbf{A}}daptive
\underline{\textbf{C}}ontinuously learning segmentation), a unified framework that integrates (i) a multi-modal backbone network trained to segment multiple types of brain lesions and handle heterogeneous sets of modalities via zero-filling and random modality dropping; (ii) 3D interactive segmentation with bounding-box and click prompts that preserves fully automatic prediction when no prompt is given; and (iii) an online adaptation mechanism combining Mid-Interaction adaptation and Post-Interaction adaptation, supervised by the network's own predictions as pseudo labels and guided by an extra Click-Centered Gaussian loss. To our knowledge, this is one of the first 3D online adaptation methods for interactive segmentation, and the first to combine handling of heterogeneous modality sets with online adaptation. Experiments across seven brain MRI datasets demonstrate that the proposed components provide complementary and synergistic benefits. The method consistently outperforms existing approaches and generalizes well across heterogeneous imaging modalities, including those unseen during training, as well as previously unseen brain pathology types. The code and a 3D Slicer plug-in  will be released at \url{https://github.com/WenTXuL/BrainIAC} upon publication.

\end{abstract}


\maketitle

\section{Introduction}
\label{sec:introduction}
Segmentation of brain lesions in Magnetic Resonance Imaging (MRI) is essential for diagnosis, treatment planning, and disease monitoring. Deep learning has achieved strong results on this task~\citep{kamnitsas2017efficient,isensee2021nnu,azad2024medical}, yet building a segmentation model that is truly effective across diverse clinical settings requires addressing several challenges.

First, automated segmentation remains unreliable in many clinical scenarios: models may miss lesion components, produce false positives from artifacts, or over/under predict lesions. Radiologists have the expertise to identify and correct such errors, however, manually editing the segmentation mask is inefficient. Interactive segmentation methods formalize this workflow by allowing clinicians to provide spatial prompts, such as clicks indicating missed or falsely detected regions, or bounding boxes localizing the area of interest. These prompts serve as extra input and help to iteratively refine predictions with the model ~\citep{sakinis2019interactive,wang2018deepigeos}. Transformer based foundation models, notably  SAM~\citep{kirillov2023segment}, have popularized this paradigm, with medical adaptations including MedSAM~\citep{ma2024segment} and
MedSAM2~\citep{ma2025medsam2}.
CNN-based foundation models, exemplified by nnInteractive~\citep{isensee2025nninteractive}, have demonstrated strong performance on a range of image segmentation tasks.
These foundation models provide strong evidence for the benefits of using interactive segmentation and training on multiple databases together. In our work, we trained a 3D interactive brain lesion segmentation model on multiple brain MRI databases. Our model retains the ability to perform fully automatic segmentation when no prompts are provided, making it well aligned with real clinical workflows in which clinician prompts may not always be available.

Second, \textbf{distribution shift remains a pervasive challenge in medical image segmentation}. Model performance often degrades when test data differ from the training distribution due to different scanners, pathologies, or acquisition protocols \textbf{and such shifts are inevitable in clinical deployment}.

Leveraging multiple datasets during training can increase data diversity and improve the model's ability to generalize across domains. However, it is impossible to cover all diversity of real-world scenarios. Interactive segmentation partially alleviates this issue: even when the test distribution differs, the model can use prompts to improve individual predictions. However, the model parameters remain frozen, consequently the same errors recur on subsequent images, requiring repeated corrections. It is therefore not sufficient to use interactions only to guide individual segmentations; the model should also learn from predictions and adapt its parameters to the new distribution. This motivates online adaptation, whereby the model is updated as images arrive sequentially, similar to online learning~\citep{hoi2021online}. Our prior OAIMS framework~\citep{xu2026you} established state-of-the-art online adaptation for interactive segmentation. It includes Post-Interaction and Mid-Interaction adaptation with a Click-Centered Gaussian (CCG) loss. However, OAIMS operates in 2D; here we extend it to 3D volumetric segmentation and demonstrate its benefits.

Third, the model should be able to handle multiple sets of modalities.
Brain MRI typically comprises a variety of imaging sequences, 
which often provide complementary diagnostic information. Hence, using multiple modalities together can  improve segmentation quality. 
In practice, different clinical protocol, pathology and/or scanner type may result in the acquisition of unique sets of modalities.
Consequently, a robust brain MRI segmentation model should be able to accommodate varying modality sets, from multiple data domains, during both training and inference. Typical segmentation models are usually trained either to segment a specific pathology, using a database acquired with a predefined set of imaging modalities determined by a protocol for the specific disease~\citep{kamnitsas2017efficient,isensee2021nnu}, or to segment multiple pathologies across different modalities while each sample contains a single modality~\citep{ma2024segment}.
Our prior work on Multi-Unet~\citep{xu2024feasibility} was the first to enable joint training and inference on heterogeneous modality sets, by zero-filling absent modalities and randomly dropping modalities during training.
Furthermore, in our previous work, (\cite{addison2025modality} we show that integrating a modality-agnostic channel allows the model to process both modalities seen during training, new modalities not seen during training, and heterogenous combinations of both. This capability of processing new modalities at inference  is absent from our previous Multi-UNet method.

Motivated by these three considerations, we introduce BrainIAC
(\underline{\textbf{Brain}} lesion
\underline{\textbf{I}}nteractive
\underline{\textbf{A}}daptive
\underline{\textbf{C}}ontinuously learning segmentation), a unified framework that addresses all of them simultaneously. BrainIAC combines a Multi-Unet backbone~\citep{xu2024feasibility} for heterogeneous modalities, 3D interactive segmentation with bounding-box and click prompts, and an online adaptation mechanism~\citep{xu2026you} that leverages user interactions to continuously adapt to new data distributions. These three components are synergistic: multi-modal joint training enriches the model's feature representations and provides a strong initialization for diverse modality combinations; interactive prompts incorporate expert knowledge to improve individual predictions; and online adaptation enables the model to learn from these interactions, progressively improving its performance on new distributions.

This work extends our prior Multi-Unet~\citep{xu2024feasibility} (MIDL 2024) and OAIMS~\citep{xu2026you} (ICLR 2026) frameworks, and demonstrates that integrating modality flexibility, interactive segmentation, and online adaptation demonstrates significant benefits.

Our key contributions are:
\begin{itemize}

\item We propose a \textbf{unified framework} for \textbf{3D interactive brain lesion segmentation} across heterogeneous modality combinations, with online adaptation and support for both fully automatic segmentation and prompt-based segmentation using bounding boxes and clicks.

     \item Our interactive model can process \textbf{multiple modality combinations}, which is closer to real-world clinical cases. Combined with an agnostic channel and online adaptation, the framework can generalize to combinations which include previously unseen modalities.
      \item We propose a \textbf{3D online adaptation} method for interactive segmentation, yielding substantial improvements under distribution shift without requiring additional annotation effort.
       \item We extensively evaluate our framework on \textbf{seven brain MRI databases} spanning diverse pathologies and modality combinations demonstrating the effectiveness of our method. Additional studies, including multi-modal fusion with nnInteractive and comparisons with other online adaptation methods, further highlight the importance of our overall design.
     
\end{itemize}

\section{Related Work}
\label{sec:related_work}
Our work lies at the intersection of three active research areas: interactive medical image segmentation, brain MRI segmentation with heterogeneous modality combinations, and online adaptation for interactive segmentation under distribution shift. We briefly review each area below.

\subsection{Interactive Medical Image Segmentation}
\label{sec:rw_interactive}

Interactive segmentation allows users to provide spatial prompts to guide model predictions. Early interactive medical image segmentation methods directly combine input images with user prompts. DeepIGeoS~\citep{wang2018deepigeos} incorporates user clicks through geodesic distance transforms, while Interactive FCNN (IFCNN)~\citep{sakinis2019interactive} encodes clicks as Gaussian guidance maps in additional input channels.

More recently, the Segment Anything Model (SAM)~\citep{kirillov2023segment} introduced a large-scale foundation model for prompt-based segmentation using a vision transformer backbone, which strongly influenced subsequent work in medical imaging. MedSAM~\citep{ma2024segment} adapts SAM to medical images through large-scale fine-tuning, while Medical SAM Adapter~\citep{wu2025medical} introduces adapter modules for efficient domain adaptation to medical data. Extensions to 3D medical imaging have also been explored. For example, SAM-Med3D~\citep{wang2025sam} develops a 3D foundation model inspired by SAM and trained on a large collection of medical datasets. MedSAM2~\citep{ma2025medsam2}, motivated by the video-capable SAM2 architecture \citep{ravi2025sam}, further extends this line of work to 3D medical images.

In parallel, nnInteractive~\citep{isensee2025nninteractive} proposes a strong 3D interactive segmentation framework built on nnU-Net~\citep{isensee2021nnu} and trained on more than 120 diverse 3D datasets spanning different medical imaging modalities and anatomies. The model supports a variety of prompt types, including points, scribbles, bounding boxes, and lasso prompts, and demonstrates strong performance through large-scale training.

Despite their success, these methods process a single modality at a time and therefore cannot leverage multiple complementary modalities available in a given brain MRI case. Moreover, their parameters remain fixed during deployment, so they cannot adapt to new data distribution shift. Our framework addresses both limitations: it supports arbitrary modality combinations within a single model and continuously adapts to the new distribution through user interactions.

\subsection{Segmentation with Heterogeneous MRI Modalities}
\label{sec:rw_multimodal}

Brain MRI commonly includes multiple imaging modalities, such as FLAIR, T1, T2, T1c, DWI, and PD, which provide complementary diagnostic information. Most existing brain MRI segmentation methods, however, are trained on a single dataset with a fixed modality set~\citep{kamnitsas2017efficient,isensee2021nnu}. This assumption is restrictive in practice, where modality availability varies across datasets, scanners, and clinical protocols. This motivates the development of a model capable of handling multiple sets of modalities during training and consequentially during inference.

Related work investigates segmentation with missing modalities, where the model is trained on a set of modalities but must remain robust when some are absent at test time.

Representative approaches include HeMIS~\citep{havaei2016hemis} and later methods based on modality synthesis or auxiliary generative components~\citep{hu2020knowledge, islam2021glioblastoma, zhou2021feature}. These methods improve flexibility at inference time, but focus on robustness to missing inputs within a fixed training setting, rather than joint training across multiple databases with different modality sets and inference with arbitrary combinations.

Our prior work~\citep{xu2024feasibility} investigate joint training on multiple brain MRI databases with heterogeneous modality sets. That framework uses the union of all modality channels across databases, zero-fills absent modalities, and applies random modality dropping during training, enabling a single model to learn from heterogeneous inputs and to exploit multiple modalities jointly when they are available for a given case. In this work, we extend this idea to the interactive segmentation setting and further investigate its interaction with online adaptation.

\subsection{Online Adaptation for Interactive Segmentation}
\label{sec:rw_adaptation}

Online adaptation methods for interactive segmentation update model parameters during deployment by utilising information provided through user interactions. The setting is closely related to online learning~\citep{hoi2021online}, where data arrive sequentially and the model is adapted as new test samples are processed.

An early representative method, IA+SA~\citep{kontogianni2020continuous}, combines independent image-level adaptation (IA) with sequence-level adaptation (SA), guided by user corrections together with cross-entropy and focal losses to update model parameters online. More recently, TSCA~\citep{atanyan2024continuous} reported further improvements in this setting. Our prior OAIMS framework~\citep{xu2026you} is, to our knowledge, the current state of the art. It uses predictions as pseudo labels and includes Post-Interaction and Mid-Interaction adaptation mechanisms, together with a Click-Centered Gaussian (CCG) loss that emphasizes the neighborhood around each user click. OAIMS consistently outperformed IA+SA and TSCA across diverse distribution-shift scenarios on 2D fundus images and brain MRI slices.

However, prior online adaptation methods for interactive segmentation have been limited to 2D. We extend online adaptation to 3D volumetric brain MRI segmentation and integrate it with a multi-modal backbone. To our knowledge, this is the first framework to combine 3D interactive online adaptation with joint handling of heterogeneous MRI modality combinations.

\section{Methods}
\label{sec:methods}

\begin{figure*}[!t]
\centering
\includegraphics[width=\textwidth]{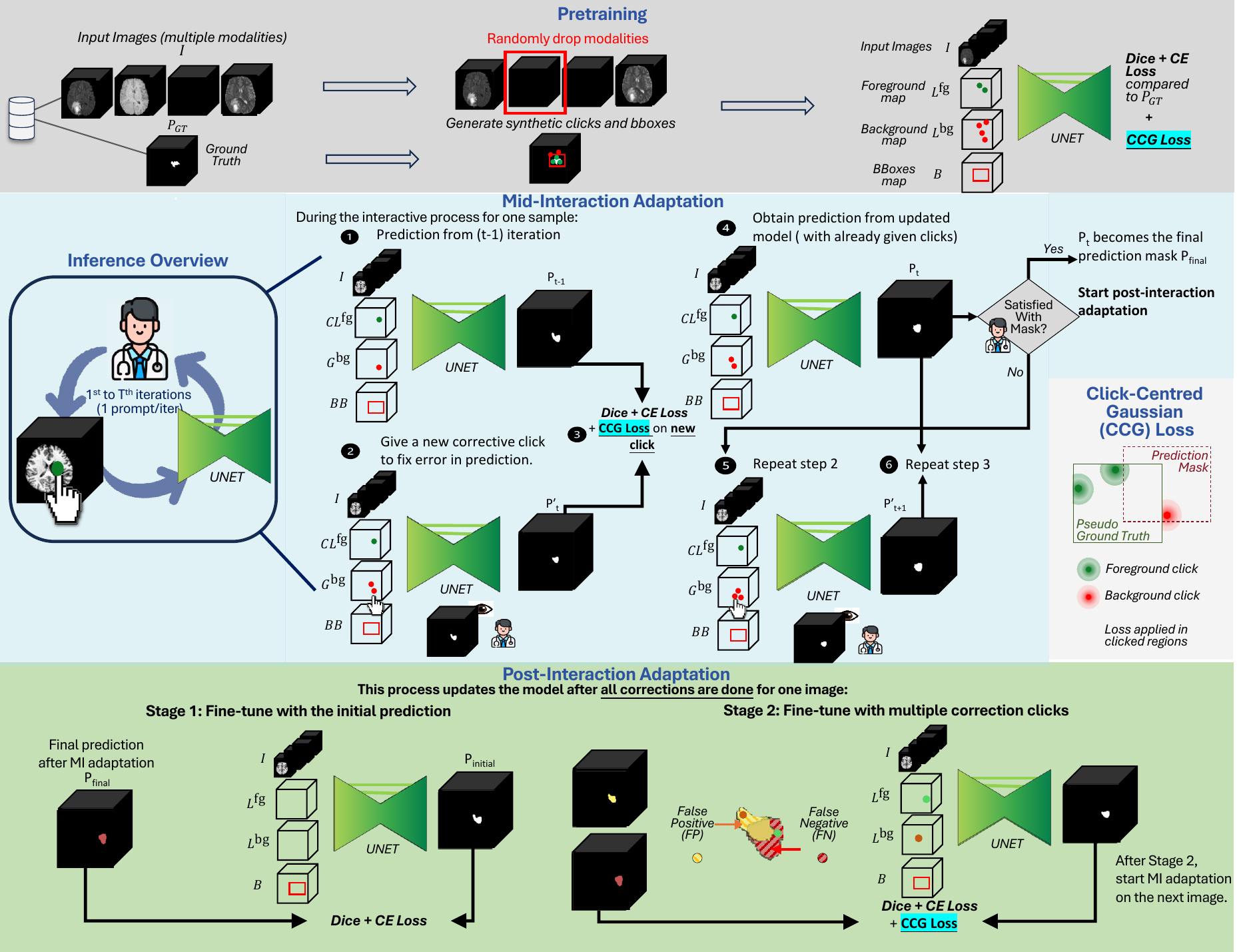}
\caption{\textbf{Overview of BrainIAC.}\textbf{Pretraining}: Simulated clicks and BBoxes are encoded in channels concatenated with the image for model training. Each modality is placed in its corresponding channel, and during training, modalities are randomly dropped to improve robustness across modality combinations. \textbf{Inference} and adaptation operate on a sequence of samples. Each sample includes multiple modalities. For each sample, the user iteratively provides one BBox for initialization and multiple clicks to correct the segmentation (one prompt per iteration), until the user is satisfied and the final prediction $P_{\text{final}}$ is obtained. In our experiments, the number of clicks $T$ is fixed for simplicity. The adaptation procedures follow our previous work~\citep{xu2026you}, applied here to 3D segmentation. In \textbf{mid-interaction adaptation}, the prediction $P_t'$ obtained with the new corrective click $l_t$ supplies pseudo supervision for the pre-correction prediction $P_{t-1}$. In \textbf{post-interaction adaptation}, the final corrected segmentation $P_{\text{final}}$ supervises two fine-tuning stages.}
\label{fig:overview}
\end{figure*}

BrainIAC is a unified framework for 3D interactive brain lesion segmentation across different sets of MRI modalities, with online adaptation serving as a key component for handling distribution shift. The framework builds on our previous work on methodology for learning from data with heterogeneous imaging modalities and the Multi-Unet~\citep{xu2024feasibility}, a method for training a modality-agnostic input channel~\citep{addison2025modality}, and the OAIMS framework for online learning from user interactions~\citep{xu2026you}. Figure~\ref{fig:overview} summarizes the overall framework. In this section, we first introduce the 3D interactive model architecture and describe how it supports diverse input modalities (Sec.~\ref{sec:multiunet}, Sec.~\ref{sec:interactive}). We next present the training and inference pipelines (Sec.~\ref{sec:training}), including  a Click-Centered Gaussian (CCG) loss (Sec.~\ref{sec:ccg}) and the 3D online adaptation mechanism (Sec.~\ref{sec:adaptation}), that allows the model to update itself during inference

\subsection{Handling Heterogeneous MRI Modality Sets}
\label{sec:multiunet}

Different databases often contain different combinations of MRI modalities, reflecting variations in clinical protocols and the pathologies under investigation. For example, glioma in BraTS is imaged with FLAIR, T1, T1c, and T2, whereas stroke in ATLAS uses only T1. To enable a single model to train jointly on databases with different modality sets $\mathbb{M}_i$ and to generalize to arbitrary modality combinations at test time, we adopt the Multi-Unet strategy from our prior work~\citep{xu2024feasibility}.

The model input has $C$ channels, where $C = |\mathbb{M}_1 \cup \ldots \cup \mathbb{M}_N|$, and is equal to the number of unique modalities across all $N$ training databases. When processing a sample from database $D_i$, only the channels corresponding to $\mathbb{M}_i$ (the set of modalities of $D_i$) contain image data; all other channels are zero-filled. This design allows a standard architecture to process inputs from any database regardless of which modalities are available, and to perform inference with arbitrary subsets of the modalities seen during training.

\textbf{Random modality dropping:} To prevent the model from associating specific modality combinations with specific pathologies, and to encourage generalization to arbitrary modality subsets at test time, we randomly drop modalities during training~\citep{xu2024feasibility}. For each training sample from database $D_i$ with $C_i = |\mathbb{M}_i|$ modalities, we sample an integer $n$ uniformly from $[0, C_i - 1]$ and randomly select $n$ modalities to replace with zeros. This also encourages the model to extract useful information from all available modalities rather than relying excessively on a single dominant modality.

\textbf{Dynamic channel addition for unseen modalities:} A limitation of the original Multi-UNet strategy is that it cannot directly handle imaging modalities absent from the training set. To address this limitation, we incorporate a modality-agnostic channel following our prior work (\cite{addison2025modality}). In this approach, a dedicated modality-agnostic channel is trained alongside modality-specific input channels using synthetically generated modality variations of both healthy and pathological tissue, augmented with information from other modalities available for the same scan. This encourages the model to learn modality-agnostic representations.

When encountering a modality not seen during training (e.g., SWI in TBI), the new modality is assigned to this modality-agnostic channel. Guided by interactive prompts, the model can immediately utilize information from the unseen modality alongside input to modality specific channels. Furthermore, through online adaptation (Sec.~\ref{sec:adaptation}), the model progressively learns to better incorporate information from the new modality during inference. This process is driven entirely by user interactions at test time, without requiring fine-tuning on the target dataset.

The synergy between the multi-modal backbone and online adaptation further enables the framework to generalize beyond the modality combinations observed during training.

\subsection{3D Interactive Segmentation Model}
\label{sec:interactive}

The backbone of BrainIAC is a 3D Residual Encoder U-Net, following the nnU-Net architecture~\citep{isensee2021nnu}. The network takes as input a tensor of size $(C + 3) \times H \times W \times D$, where $C$ denotes the number of imaging modality channels, and the additional three channels encode user prompts: (1) a bounding-box mask, (2) a foreground click guidance map, and (3) a background click guidance map. The network outputs a single-channel segmentation map with the same spatial dimensions as the input. 
The final prediction is a binary mask, where lesion voxels are assigned label 1 and background voxels are assigned label 0.

For the prompt channels, each foreground or background click sets the corresponding voxel in the foreground or background click guidance map channel to 1. For the bounding-box prompt, all voxels inside the box are set to 1 in the bounding-box channel. The method handles 2D bounding boxes as they are easier to "draw" for a human user. In the mask, the 2D box is extended (duplicated) to the 2 adjacent slices along the depth dimension, for a total area of $(w, h, 3)$, where $w$ and $h$ are width and height. This allows the 3D convolutional kernels of the model to better capture the box, for improved spatial guidance in 3D.

When no prompt is provided, the corresponding prompt channels are zero-filled, allowing the model to operate in a fully automatic mode when all three prompt channels are empty. This provides a native formulation for automatic segmentation, in contrast to methods such as SAM-based \citep{kirillov2023segment} approaches that require generated prompts even for automatic segmentation.

\subsection{Training}
\label{sec:training}

The model is trained jointly on multiple databases with the Dice cross-entropy (DCE) loss combined with the Click-Centered Gaussian (CCG) loss (detailed in Sec.~\ref{sec:ccg}):
\begin{equation}
\mathcal{L}_{\text{train}} = \mathcal{L}_{\text{DCE}} + \beta \, \mathcal{L}_{\text{CCG}},
\label{eq:train_loss}
\end{equation}
where $\mathcal{L}_{\text{DCE}} = \mathcal{L}_{\text{Dice}} + \mathcal{L}_{\text{CE}}$, and $\mathcal{L}_{\text{Dice}}$ and $\mathcal{L}_{\text{CE}}$ denote the Dice loss~\citep{milletari2016vnet} and cross-entropy loss, respectively, and $\beta$ is a weighting hyperparameter. The CCG loss is applied whenever clicks are used as prompts, reinforcing the model's responsiveness to user corrections. Our prior work~\citep{xu2026you} demonstrated the effectiveness of this loss.

\textbf{Prompt sampling strategy.} For each training sample, a bounding box is included with 70\% probability, and clicks are included independently with 70\% probability. This yields four possible prompt configurations: both bounding box and clicks (49\%), bounding box only (21\%), clicks only (21\%), and no prompts (9\%). This design serves two purposes: it enables the model to handle any prompt combination at test time, and it preserves fully automatic segmentation ability---unlike nnInteractive~\citep{isensee2025nninteractive} and MedSAM2~\citep{ma2025medsam2}, which require prompts at inference.

For click prompts, we simulate correction clicks to mimic user behavior. The model first produces an initial prediction (using the bounding box if present) without any clicks. This prediction is compared to the ground truth to identify false negatives (missed foreground) and false positives (incorrectly predicted foreground). To ensure the model is sufficiently exposed to both types of clicks, we randomly draw clicks, with equal probability, entirely from false negatives, entirely from false positives, or from a mixture of both. To keep training efficient, we do not adopt a more complex strategy. The number of clicks $L$ is sampled uniformly from $[1, 10]$ at each training iteration.

For bounding-box prompts, the bounding box is drawn to cover foreground components, with random padding of 0--10\%. During training, we randomly select a subset of foreground connected components (up to 6). For each component, we generate 1--4 2D bounding boxes. Each box is sampled along a random orthogonal plane ($xy$, $yz$, or $xz$), providing diverse bounding-box configurations.

\subsection{Click-Centered Gaussian (CCG) Loss}
\label{sec:ccg}

An effective interactive model should react to user clicks and update the segmentation of surrounding region accordingly. For this, we employ the Click-Centered Gaussian (CCG) loss, introduced in our previous study \citep{xu2026you}. Optimizing model parameters with the CCG loss improves the model's ability to react to user clicks and make appropriate corrections of its prediction, by penalizing incorrect predictions near each click, weighted by a Gaussian kernel.  The CCG loss is used both during training and during online adaptation. Here, we adapt the formulation of the CCG loss to be amenable for 3D models, extending the original 2D version. 
Let $\mathbf{v}_{l}=(i', j', k')$ denote a voxel where a click $l$ was given, with class label $y_{l} \in \{0,1\}$, and let $d_{l}(\mathbf{v})=\|\mathbf{v}-\mathbf{v}_{l}\|_2$ be the Euclidean distance from voxel $\mathbf{v}=(i,j,k)$ to the click. We define a 3D Gaussian weight:
\begin{equation}
G_{l}(\mathbf{v}) =
\begin{cases}
\exp\!\bigl(-{d_{l}(\mathbf{v})^2}/{2\sigma^2}\bigr), & \text{if } d_{l}(\mathbf{v}) \leq 3\sigma, \\
0, & \text{otherwise,}
\end{cases}
\label{eq:gaussian}
\end{equation}
and a class-limited indicator:
\begin{equation}
I_{l}(\mathbf{v}) =
\begin{cases}
1, & \text{if } P(\mathbf{v}) = y_{l}, \\
0, & \text{otherwise}.
\end{cases}
\label{eq:indicator}
\end{equation}
Here, $P$ denotes the ground-truth mask during training, or the pseudo ground-truth during adaptation. 

Let $w_{l}(\mathbf{v}) = G_{l}(\mathbf{v})\, I_{l}(\mathbf{v})$. The CCG loss is:
\begin{equation}
\mathcal{L}_{\text{CCG}} =
\frac{1}{|L|\,N}
\!\sum_{l,\,\mathbf{v}}
w_{l}(\mathbf{v})\;
\mathrm{CE}\!\bigl(\hat{P}(\mathbf{v}), P(\mathbf{v})\bigr),
\label{eq:ccg}
\end{equation}
where $N\!=\!H\!\times\!W\!\times\!D$ is the number of voxels, $\hat{P}$ is the model prediction, $L$ is the set of clicks, and $\mathrm{CE}$ denotes cross-entropy. The weight $w_{l}$ combines the Gaussian proximity and class indicator, so the penalty is applied only to voxels near the click that share the same class. For example, for a foreground click, the loss acts only on nearby foreground voxels.

\subsection{Inference and Online Adaptation}
\label{sec:adaptation}

During inference, the model processes images sequentially and responds to the clinician after each new prompt. Our method supports multiple different interactive strategies; however, here we define a default strategy to allow for simple comparison. The interaction strategy is as follows: for each image, the model first produces an initial automated prediction, optionally using a clinician-provided bounding box for localization. The clinician then reviews the result and provides correction clicks. After each new click, the model generates an updated prediction using all the prompts accumulated so far for the current image and displays the result, allowing the clinician to decide whether further corrections are needed. By default, the bounding box is placed on the $xy$ plane over the largest foreground component, with a randomly chosen $z$ index, and each correction click is randomly generated within the largest connected component of the erroneous region.
For simplicity in our experiments, we assume that the clinician stops after $T$ interactions on each image, although in real-world practice the number of interactions may vary between cases.

Instead of freezing the model at test time, we leverage the clinician's interactions and corrected predictions to perform online adaptation. The adaptation consists of two complementary mechanisms. \emph{Mid-Interaction} (MI) adaptation updates the model after each correction click, improving both the segmentation and subsequent predictions for the current image. \emph{Post-Interaction} (PI) adaptation further updates the model after the clinician completes all interactions for an image, improving future predictions on later images. This framework extends our previous OAIMS method~\citep{xu2026you} from 2D to 3D interactive segmentation. 
\textbf{In principle,} MI adaptation can also be applied after introducing a bounding box. However, in our method we primarily focus on correction clicks, which are the most direct and common form of iterative user feedback.

\subsubsection{Mid-Interaction (MI) Adaptation}

At interaction step $t$, let $P_{t-1} = f(I, L_{t-1}; \theta)$ and $P_t' = f(I, L_t; \theta)$ denote the predictions before and after adding the corrective click $l_t$, respectively, where $L_t = L_{t-1} \cup \{l_t\}$. Both predictions use the same parameters $\theta$. We use $P_t'$ as a pseudo target for $P_{t-1}$:
\begin{equation}
\mathcal{L}_{\text{MI}} =
\mathcal{L}_{\text{DCE}}(P_{t-1}, P_t')
+ \beta \, \mathcal{L}_{\text{CCG}}(P_{t-1}, P_t', l_t).
\label{eq:mi}
\end{equation}
The CCG loss is only applied to the latest click $l_t$ to emphasize supervision near the user's correction. After updating the parameters to $\theta_{\text{new}}$, we compute $P_t = f(I, L_t; \theta_{\text{new}})$ and display it to the user. This output becomes the pre-correction prediction for the next interaction, and the updated parameters are retained for subsequent cases. For simplicity, we omit the localization bounding-box prompt from the equations, but it is always used as part of the input.

\subsubsection{Post-Interaction (PI) Adaptation}

After $T$ interactions, the final corrected segmentation $P_{\text{final}} = P_T$ provides pseudo supervision for two adaptation stages. As in OAIMS \citep{xu2026you}, this assumes that the user's corrections yield a sufficiently high quality target mask to serve as a supervision signal.

\textbf{Stage 1 -- Fine-tuning for initial prediction.}
We obtain $P_{\text{initial}} = f(I; \theta)$ using the original initialization setting: no prompts or the initial bounding box, without correction clicks. With $P_{\text{final}}$ as the target, we perform one gradient descent update per image using $\mathcal{L}_{\text{DCE}}$. 

\textbf{Stage 2 -- Fine-tuning for correction clicks.}
Following OAIMS, we generate a fresh click set $\hat{L}$ rather than reuse the clicks that produced $P_{\text{final}}$, which could yield weak or trivial updates. We compare the Stage~1 prediction $P_{\text{initial}}$ with $P_{\text{final}}$ to identify false-positive and false-negative regions, placing one click in each erroneous connected component, up to $T$ clicks in total. 

The resulting prediction $\hat{P} = f(I, \hat{L}; \theta)$ is optimized against $P_{\text{final}}$ using the combined DCE and CCG losses in Eq.~\ref{eq:train_loss}. If a bounding box was used in the original interaction, it is retained as an input in Stage~2, although it is omitted from the notation above for simplicity.

\section{Experiments and Results}
\label{sec:experiments}

\subsection{Datasets}
\label{sec:datasets}

The proposed method was assessed across seven brain MRI datasets spanning multiple pathologies and MRI modality combinations, thereby capturing substantial clinical and imaging heterogeneity: \textbf{BraTS 2016}~\citep{bakas2017brats}: brain tumour (FLAIR, T1, T1c, T2; 444/40 train/test);
\textbf{ATLAS}~\citep{liew2022atlas}: stroke (T1; 459/195 train/test);
\textbf{MSSEG}~\citep{commowick2018msseg}: multiple sclerosis (FLAIR, T1, T1c, T2, PD; 37/16 train/test);
\textbf{WMH}~\citep{kuijf2019wmh}: white matter hyperintensities (FLAIR, T1; 42/18 train/test);
\textbf{TBI}: database of patients with traumatic brain injury, collected at Addenbrookes Hospital, Cambridge, UK (FLAIR, T1, T2, SWI; 156/125 train/test);
\textbf{VES-SEG}~\citep{shapey2021segmentation,datashapey2021schwannoma}: vestibular schwannoma (T1c, T2; 242 test-only cases);
\textbf{ISLES}~\citep{maier2017isles}: stroke (FLAIR, T1, T2, DWI; 28 test-only cases).
TBI is the only database collected at our institution; all other databases are publicly available. All 3D scans were resampled to an isotropic resolution of $1 \times 1 \times 1$ mm and standardized to a uniform size of $192 \times 192 \times 192$ voxels. Intensity values were subsequently normalized using Z-score normalization.

We consider three training configurations: (i)~\textbf{4-DB$_{\text{MSSEG}}$}, in which the model was jointly trained on BraTS, ATLAS, WMH, and TBI, and evaluated on the held-out MSSEG dataset;
(ii)~\textbf{4-DB$_{\text{TBI}}$}, in which the model was jointly trained on BraTS, ATLAS, WMH, and MSSEG, and evaluated on the held-out TBI dataset; and
(iii)~\textbf{5-DB}, in which the model was jointly trained on all five datasets used in the main training pool (BraTS, ATLAS, WMH, MSSEG, and TBI), and evaluated on VES-SEG and ISLES.

For the evaluation of a held-out dataset, we tested a combined set of samples from the original training and test splits, since the entire dataset was excluded from model training and used only for out-of-distribution evaluation and online adaptation.

\begin{figure*}[!t]
\centering
\includegraphics[width=\textwidth]{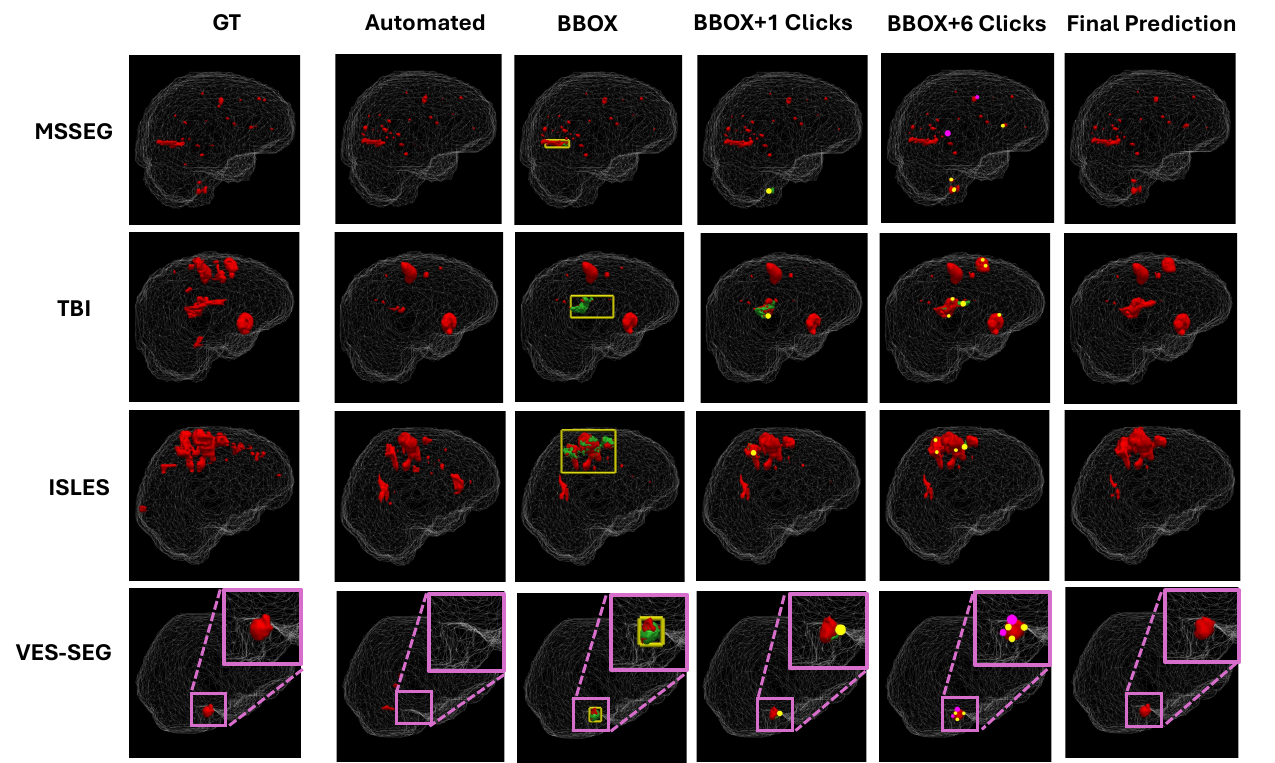}
\caption{Illustration of how the predicted segmentation from BrainIAC evolves as the BBox and interaction clicks are provided.The examples are, from top to bottom, MSSEG, TBI, ISLES and VES-SEG. From left to right: ground truth, automated prediction, BBox only, BBox+1 click, BBox+6 click, and the final prediction. Positive clicks are shown in yellow, negative clicks in purple, and clicks are accumulated across interactions. Regions in \textbf{\textcolor{red}{red}} denote the current prediction, while regions highlighted in \textbf{\textcolor{ForestGreen}{green}} indicate improvements resulting from the current prompt. The results demonstrate progressive refinement of the segmentation as more prompts are provided. }
\label{fig:overview_diagram}
\end{figure*}

\subsection{Implementation Details}
\label{sec:impl}

The input to the model consists of $C$ imaging modality channels and 3 prompt channels, corresponding to one bounding-box channel and two click guidance channels (foreground and background). The number of modality channels $C$ is determined by the number of unique modalities across the training datasets. In the \textbf{5-DB} configuration, the union of modalities is \{FLAIR, T1, T1c, T2, PD, SWI\}, resulting in $C = 6$ and 9 total input channels. In the \textbf{4-DB} configurations, one dataset is held out, reducing the modality union accordingly: when MSSEG is held out, PD is absent ($C = 5$, 8 total channels), and when TBI is held out, SWI is absent ($C = 5$, 8 total channels). 
All volumes are processed as full volumes. The model is trained with batch size 1 and a learning rate of $10^{-3}$. Up to $K = 10$ clicks are simulated in each training iteration. The loss weight is set to $\beta = 10000$, and the CCG Gaussian uses $\sigma = 2$.  $\beta$ is large because the CCG loss is normalized by the full-volume voxel count $N = 192^3$, while the Gaussian is nonzero only within a small region. For online adaptation, both Mid-Interaction (MI) and Post-Interaction (PI) updates use the Adam optimizer with a learning rate of $5 \times 10^{-5}$, performing a single gradient descent step at each update.

\subsection{Evaluation Protocol}
\label{sec:eval}

By default, we evaluate the model under multiple interaction settings:
\textbf{No Prompt} (fully automatic, with no bounding box or clicks),
\textbf{BBox} (bounding box provided on the largest foreground components, with no clicks), and
\textbf{BBox+$k$} (bounding box plus $k$ iterative correction clicks, where $k \in \{1,6\}$).

For online adaptation experiments, metrics are computed over the full sequential adaptation process. Test images are processed one by one in sequence. For each image, the model undergoes Mid-Interaction (MI) adaptation after each click interaction and Post-Interaction (PI) adaptation after all interactions are completed. The Dice score is recorded after each interaction step, and we report the average Dice over all test images. Unless otherwise specified, each image receives one bounding box and 6 correction clicks. 
We also conduct experiments that aim to reach a target Dice score rather than performing a fixed number of clicks.

\subsection{Baselines}
\label{sec:baselines}

We compare against two categories of baselines. \emph{3D Interactive Segmentation Foundation Models:}
\textbf{nnInteractive}~\citep{isensee2025nninteractive}: an interactive segmentation framework trained on over 120 datasets, which processes a single modality as input and does not perform online adaptation.
\textbf{MedSAM2}~\citep{ma2025medsam2}: a 3D adaptation of SAM2~\citep{ravi2025sam} fine-tuned on large-scale medical data, supporting bounding-box prompts. This method also processes a single modality as input and does not perform online adaptation.

\emph{Online Adaptation Methods:}
\textbf{IA+SA}~\citep{kontogianni2020continuous}: the first online adaptation method for interactive segmentation, combining image-level and sequence-level adaptation.
\textbf{TSCA}~\citep{atanyan2024continuous}: the previous state-of-the-art method for online adaptation in interactive segmentation.

For fair comparison, IA+SA and TSCA use the same backbone architecture as our method.  As a result, they can also handle multiple modality combinations, allowing the comparison to focus on the online adaptation strategy.
The original IA+SA and TSCA were applied to 2D images; we extend them to 3D. The backbone architecture alone, without online adaptation, corresponds to the \textbf{No Online} results in the present experiments.

\subsection{Evaluation on Unseen Pathologies}
\label{sec:results_ood}
We primarily evaluate our method on datasets unseen during training, as this setting is more challenging and reflects real-world deployment. In these experiments, the target pathology and dataset are excluded entirely from training.

\subsubsection{MSSEG -- Multiple Sclerosis}
\label{sec:MSSEG}

Using the \textbf{4-DB$_{\text{MSSEG}}$} configuration, and evaluated on the held-out MSSEG dataset. None of the compared methods in this setting, including the foundation models, are trained on MSSEG.

Results are shown in Table~\ref{tab:msseg_ood}. BrainIAC  can operate in a fully automatic mode (\textbf{No Prompt}), whereas MedSAM2 and nnInteractive require user prompts. Even without online adaptation, BrainIAC already performs strongly on this unseen pathology. Using only FLAIR, BrainIAC outperforms nnInteractive and MedSAM2 across all comparable settings. Following the bounding-box prompting protocol described in the MedSAM2 paper, we evaluate MedSAM2 with bounding boxes and do not include it in the point-based comparisons. Adding T1 further improves the performance of BrainIAC, demonstrating the benefit of jointly exploiting multiple modalities.

Online adaptation yields further consistent improvements in all modality settings. In particular, our online adaptation version outperforms both IA+SA and TSCA, the two  online adaptation baselines, which is consistent with the findings of our previous work on 2D online adaptation for interactive segmentation. Under online adaptation, adding T1 still improves the result, but not as much as without online adaptation.

Each subject in MSSEG may contain many small and scattered lesions. In such cases, a single bounding box provides only limited information. This helps explain the weaker performance of MedSAM2, which relies on bounding-box prompting. To investigate this further, we evaluated the interactive segmentation methods under a \textbf{Top-5 BBox} prompt setting, in which the five largest connected lesion components are each given a bounding box at initialization (lower block of Table~\ref{tab:msseg_ood}). For a fair comparison with the baseline, we use only the FLAIR modality. All methods benefit from the additional BBoxes prior to clicks: nnInteractive improves from 34.6 to 42.8 at the BBox stage, however, a slight decrease from 52.0 to 50.5 at BBox+6 (additional clicks appear to help less when the model already has multi-box initialization). MedSAM2 also improves under 5 BBoxes. BrainIAC improves from 66.2 to 72.8 (single BBox) and from 69.8 to 75.2 (Top-5 BBox) for BBox and BBox+6 respectively. Even with five boxes, MedSAM2 and nnInteractive results remain substantially below our method, and providing one box per lesion is impractical in realistic clinical scenarios.

In addition, we also test the setting without any bounding boxes, since click-based interaction is more flexible. We report click-only experiments on MSSEG (\textbf{Pts-1}, \textbf{Pts-6} columns). Compared with nnInteractive, our method performs substantially better both with and without online adaptation. For fair comparison, the first click is randomly generated on the biggest lesion for all of the click-only experiments, since nnInteractive requires an initial click to start segmentation, even though our method does not.

\begin{table*}[!t]
\centering
\caption{Results on MSSEG in the held-out setting, where MSSEG is excluded from training. Dice score (\%). The main interaction setting starts from automatic prediction and is followed by bounding-box initialization and iterative correction (\textbf{No Prompt}, \textbf{BBox}, \textbf{BBox+1}, \textbf{BBox+6}). Click-only results (\textbf{Pts-1}, \textbf{Pts-6}) are also reported for comparison with methods that require click-based initialization. ``BrainIAC (ours) - No Online'' refers to the backbone 3D interactive model which does not include online adaptation, while ``BrainIAC (ours) - Online'' includes online adaptation. The lower block reports the \textbf{Top-5 BBox} setting, in which the five largest lesion components are each given a bounding box at initialization. Best result in each column is shown in \textbf{bold}.}
\label{tab:msseg_ood}
\footnotesize
\begin{tabular}{@{}llcccccc@{}}
\toprule
& & \multicolumn{4}{c}{Main settings} & \multicolumn{2}{c}{Click-only} \\
\cmidrule(lr){3-6} \cmidrule(l){7-8}
Method & Modalities & No Prompt & BBox & BBox+1 & BBox+6 & Pts-1 & Pts-6 \\
\midrule
BrainIAC (ours) - No Online & FLAIR       & 53.6 & 57.3 & 60.7 & 68.5 & 56.2 & 67.5 \\
BrainIAC (ours) - Online      & FLAIR       & 64.7 & 66.2 & 68.1 & 72.8 & 66.3 & 72.5 \\
BrainIAC (ours) - No Online & FLAIR, T1   & 55.0 & 59.1 & 62.0 & 69.6 & 57.3 & 68.2 \\
BrainIAC (ours) - Online    & FLAIR, T1   & \textbf{66.1} & \textbf{66.9} & \textbf{68.7} & \textbf{73.3} & \textbf{67.2} & \textbf{72.9} \\
\midrule
TSCA     & FLAIR      & 60.8 & 63.1 & 66.0 & 72.0 & 63.8 & 71.5 \\
TSCA     & FLAIR, T1  & 62.4 & 63.8 & 66.2 & 71.8 & 64.1 & 72.2 \\
IA+SA    & FLAIR      & 58.4 & 61.5 & 64.8 & 71.1 & 61.8 & 70.1 \\
IA+SA    & FLAIR, T1  & 60.0 & 62.8 & 65.8 & 71.9 & 63.0 & 70.8 \\
\midrule
nnInter. & FLAIR & -- & 34.6 & 41.8 & 52.0 & 16.8 & 47.0 \\
MedSAM2  & FLAIR & -- & 40.9 & -- & -- & -- & -- \\
\midrule
\multicolumn{8}{@{}l}{\textit{Top-5 BBox prompting (5 largest-component bounding boxes provided 
at initialization)}} \\
\midrule
BrainIAC (ours) - No Online (5 BBox) & FLAIR & -- & 65.0 & 66.9 & 71.9 & -- & -- \\
BrainIAC (ours) - Online (5 BBox)    & FLAIR & -- & \textbf{69.8} & \textbf{71.3} & \textbf{75.2} & -- & -- \\
nnInter. (5 BBox) & FLAIR & -- & 42.8 & 44.5 & 50.5 & -- & -- \\
MedSAM2 (5 BBox)  & FLAIR & -- & 46.8
& -- & -- & -- & -- \\
\bottomrule
\end{tabular}
\end{table*}

\subsubsection{TBI -- Traumatic Brain Injury}
TBI is unique as different lesion components are visible on FLAIR and SWI respectively. Consequentially, clinicians annotated each case with separate segmentation masks for FLAIR and SWI. We therefore report two evaluation settings: one restricted to FLAIR-visible lesions, which provides the fairest comparison for methods that use only FLAIR as an input, and one using the merged annotations that include all lesion components (SWI and FLAIR). The second one is useful for showing the model's ability to utilise additional diagnostic information from modalities unseen during training.

In this section, we first focus on FLAIR-visible lesions (Table~\ref{tab:tbi}), with all experiments performed using the \textbf{4-DB$_{\text{TBI}}$} configuration. None of the compared methods including the foundation models are trained on TBI. In this setting, our model without online adaptation is weaker than nnInteractive and MedSAM2. This is not unexpected, since nnInteractive and MedSAM2 are pretrained on large datasets and therefore, may achieve stronger initial performance on some target datasets. However, once online adaptation is enabled, our method surpasses nnInteractive and achieves the best performance when using FLAIR alone or FLAIR together with T1 and T2. This shows that in some cases where the initial segmentation is not as strong, online adaptation can still effectively adapt the model to a new target distribution and exceed the performance of strong pretrained models. In previous experiments, we already showed cases in which our pretraining was stronger; here, we highlight the complementary case in which the initial model is weaker, but online adaptation still achieves the best final performance.

The benefit of online adaptation is even more pronounced on TBI than on MSSEG. The same overall trend observed on MSSEG is also present here: our continual version consistently outperforms the previous online adaptation baselines TSCA and IA+SA. We also observe that adding modalities improves performance both with and without online adaptation.

In addition, we report results evaluated against labels that include all lesion segments, including those visible only on SWI. However, SWI is not used as an input modality in these experiments, and consequently the overall Dice scores are lower. One advantage of BrainIAC is that it is designed to support different modality combinations within a single model. However, a modality not seen during training cannot be directly exploited. In the next section (Sec.~\ref{sec:results_multichannel}) we demonstrate that an additional \textbf{modality-agnostic channel} and interactive segmentation enable our framework to learn to use such unseen modalities.

\begin{table*}[!t]
\centering
\caption{Results on TBI. Dice score (\%). \emph{FLAIR-visible lesions}: only lesions visible on FLAIR are evaluated, providing the fairest comparison since all methods receive FLAIR as input. \emph{All lesions}: lesions visible on SWI are also included, although SWI is not provided as an input modality, making the task more challenging. Best result in each column (within each setting) is shown in \textbf{bold}.}
\label{tab:tbi}
\footnotesize
\begin{tabular}{@{}llcccc@{}}
\toprule
Method & Modalities & No Prompt & BBox & BBox+1 & BBox+6 \\
\midrule
\multicolumn{6}{@{}l}{\emph{FLAIR-visible lesions:}} \\
\midrule
BrainIAC (ours) -- No Online & FLAIR          & 37.9 & 39.5 & 41.1 & 47.9 \\
BrainIAC (ours) -- Online    & FLAIR          & 44.3 & 50.5 & 54.3 & 62.2 \\
BrainIAC (ours) -- No Online & FLAIR, T1, T2  & 41.5 & 43.7 & 46.1 & 52.4 \\
BrainIAC (ours) -- Online   & FLAIR, T1, T2  & \textbf{46.6} & \textbf{52.6} & \textbf{55.8} & \textbf{62.7} \\
\midrule
TSCA     & FLAIR          & 37.3 & 41.3 & 45.2 & 55.3 \\
TSCA     & FLAIR, T1, T2  & 41.7 & 45.1 & 48.5 & 58.1 \\
IA+SA    & FLAIR          & 37.8 & 40.2 & 43.4 & 51.8 \\
IA+SA    & FLAIR, T1, T2  & 41.8 & 44.6 & 48.1 & 57.0 \\
\midrule{}
nnInter. & FLAIR          & -- & 43.7 & 51.1 & 60.1 \\
MedSAM2  & FLAIR          & -- & 47.9 & -- & -- \\
\midrule
\multicolumn{6}{@{}l}{\emph{All lesions (including SWI-visible lesions):}} \\
\midrule
BrainIAC (ours) -- No Online & FLAIR          & 36.7 & 36.8 & 38.5 & 44.0 \\
BrainIAC (ours) -- Online   & FLAIR          & 42.0 & 46.9 & 50.6 & 58.0 \\
BrainIAC (ours) -- No Online & FLAIR, T1, T2  & 38.4 & 38.5 & 40.5 & 46.4 \\
BrainIAC (ours) -- Online   & FLAIR, T1, T2  & \textbf{43.6} & \textbf{49.2} & \textbf{52.2} & \textbf{58.7} \\
\bottomrule
\end{tabular}
\end{table*}
\subsection{Evaluation on Unseen Modalities}
\label{sec:results_multichannel}
A key advantage of BrainIAC is its ability to handle arbitrary combinations of imaging modalities within a single model. However, the original Multi-UNet architecture cannot directly process modalities absent from the training set. To address this limitation, We investigate two separate strategies: randomly initializing an additional channel and utilizing a pretrained modality-agnostic channel.

\textbf{Additional channel:} a new input channel is appended to the pretrained model with randomly initialized weights, and the SWI modality is assigned to this new channel. Although online adaptation is applied during interactive segmentation, this strategy provides limited improvement compared to the model without the additional channel. A randomly initialized channel creates a large distribution mismatch, preventing the model from learning meaningful representations of the unseen modality from limited user interactions. As a result, it provides little benefit during inference. Motivated by this limitation, we instead utilise a modality-agnostic channel.

\noindent\textbf{Pretrained Agnostic Channel}: We adapted our model using the modality-agnostic path framework we previously proposed (\cite{addison2025modality}).
Additionally, a Bézier-curve-based non-linear intensity remapping augmentation was employed, similar to Zhou et al. (2022). The model was initialised from a pretrained model (\textbf{4-DB$_{\text{TBI}}$}) while the modality-agnostic channel was randomly initialised prior to training. As shown in Table~\ref{tab:multichannel}, incorporating SWI through this channel improves performance after continual learning achieving a BBox+6 Dice score of 61.0 compared to 58.7 from our original model which cannot process SWI. Furthermore, when the agnostic channel is zero-filled, performance (58.6) remains comparable to that of our original model with FLAIR only, demonstrating that the improvements observed with SWI input are attributable to the additional diagnostic information provided by the new modality.

\begin{figure*}[!t]
\centering
\includegraphics[width=\textwidth]{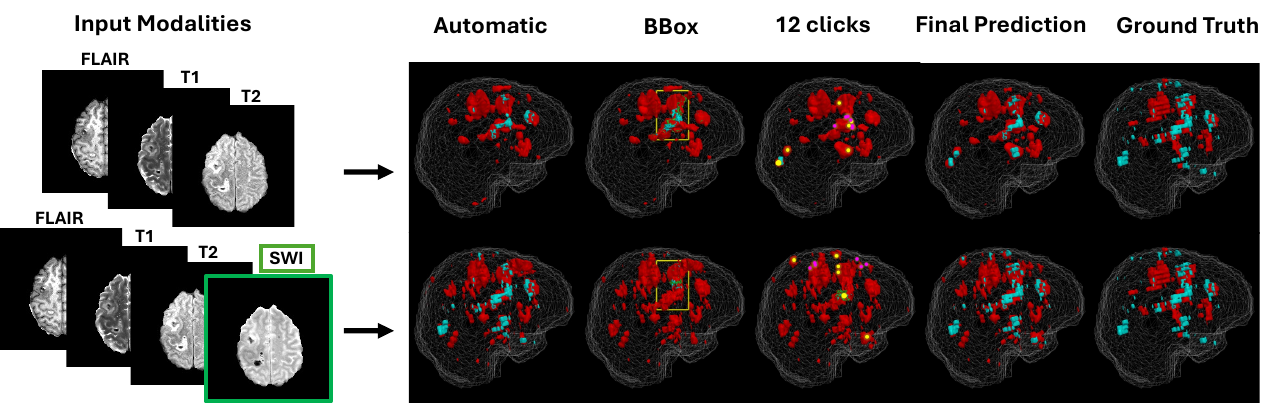}
\caption{TBI sample and three-dimensional visualisation of the progressive refinement of the predicted segmentation with a BBox and 12 subsequent interaction clicks, which alternate between the largest global segmentation error and the largest remaining unsegmented SWI lesion component. The top row shows results obtained with our model using T1, T2, and FLAIR as inputs during online adaptation. The bottom row shows the corresponding results when SWI (highlighted as \textbf{\textcolor{Aquamarine}{blue}} in ground truth) is additionally incorporated through an agnostic input channel. Positive clicks  in yellow, negative clicks in purple, clicks accumulated across interactions. \textbf{\textcolor{red}{Red}} denotes current prediction. \textcolor{Aquamarine}{Blue} highlights correctly predicted SWI components. \textbf{\textcolor{ForestGreen}{Green}} highlights correction of predicted FLAIR label at current interaction step. This iterative interaction strategy progressively refines the segmentation and improves lesion segmentation.}
\label{fig:TBI_agnostic_channel_images}
\end{figure*}

We further investigated \textbf{alternative click strategies} for continual learning to better exploit the SWI modality, as summarized in Table~\ref{tab:TBI_unused modality}. Specifically, we evaluated a strategy that alternates user clicks between the largest segmentation error and largest undetected SWI lesion component, for a total of 12 clicks (a representative example can be seen in figure \ref{fig:TBI_agnostic_channel_images}). This resulted in a much higher percentage of SWI connect components being detected when using the agnostic channel (68.9\%) compared to our original set up with no place to process SWI (53.1\%), however, we still click where the largest undetected SWI lesions would be using our SWI ground truth. This lesion-type-specific clicking strategy improves the detection rate of SWI lesion components compared with our standard click strategy: BBox + 6 largest error clicks for the agnostic channel with SWI (49.8\%) and ours without SWI (38.6\%) and is more representative of how a clinician would use this tool in practice. The randomly initialized channel provided limited improvements under both click strategies.

\textbf{Baseline - process extra modality:} A natural question is whether single-modality methods such as nnInteractive could benefit from multiple modalities by combining predictions from separate inference passes, without the added effort of learning joint multi-modal representations. Table~\ref{tab:multichannel} compares two such strategies: \textbf{UNION}, which takes the union of lesion predictions from all modalities, and \textbf{LOGITS}, which averages prediction logits across modalities before applying the final softmax. Both strategies show inconsistent or degraded performance as additional modalities are introduced. In particular, UNION degrades monotonically: its BBox+6 Dice decreases from 54.8 using FLAIR alone to 46.8 when all channels are combined. This suggests that independently processed modalities often produce conflicting rather than complementary predictions. LOGITS performs better in some cases, for example on FLAIR+SWI at BBox+6, indicating that it can capture some useful complementary information across modalities. However, its performance remains unstable overall and does not achieve the same level of performance as our agnostic channel method.

\begin{table*}[!t]
\centering

\caption{Evaluation on unseen modalities for TBI. Dice score (\%). This table reports results evaluated with labels containing both lesions visible on FLAIR and lesions visible on SWI, while using SWI as an extra input to the model. It also reports results from using multiple nnInteractive models for different modalities and combining the result. Best result in each section is shown in \textbf{bold}. (Extra Ch.: Extra channel initiated, Agn.: Agnostic channel initiated)}
\label{tab:multichannel}
\footnotesize
\setlength{\tabcolsep}{6pt}
\begin{tabular}{@{}llcccc@{}}
\toprule
Method & Channels & Auto & BBox & BBox+1 & BBox+6 \\
\midrule

BrainIAC (ours) -- Online    & FLAIR                  & 42.0 & 46.9 & 50.6 & 58.0 \\
BrainIAC (ours) -- No Online & FLAIR, T1, T2  & 38.4 & 38.5 & 40.5 & 46.4 \\
BrainIAC (ours) -- Online     & FLAIR, T1, T2          & 43.6 & 49.2 & 52.2 & 58.7 \\

BrainIAC (ours) + Extra Ch. -- Online     & FLAIR,SWI        & 43.7 & 44.0 &49.0  & 57.7   \\
BrainIAC (ours) + Extra Ch. -- Online     & FLAIR,T1,T2,SWI & 44.1& 49.2& 52.5& 58.7 \\
BrainIAC (ours) + Agn. -- Online     & FLAIR,SWI       & 44.6 & 49.4 & \textbf{53.8} & \textbf{61.0}  \\
BrainIAC (ours) + Agn. -- No Online      & FLAIR,T1,T2,SWI         & 37.0 & 40.1& 43.8 & 50.6 \\
BrainIAC (ours) + Agn. -- Online    & FLAIR,T1,T2,\textit{Zero}         & 44.6& 48.9 & 52.6 & 58.6 \\
BrainIAC (ours) + Agn. -- Online    & FLAIR,T1,T2,SWI         & \textbf{46.3} & \textbf{50.4} & 53.5 & \textbf{61.0} \\
\midrule
nnInter. (UNION)      & FLAIR                  & -- & 42.3 & 48.6 & 54.8 \\

nnInter. (UNION)      & FLAIR,SWI             & -- & 40.2 & 43.1 & 50.2 \\
nnInter. (UNION)      & FLAIR,T1,T2         & -- & 39.9 & 44.1 & 48.3 \\
nnInter. (UNION)      & FLAIR,T1,T2,SWI                       & -- & 38.3 & 41.5 & 46.8 \\
\midrule
nnInter. (LOGITS)     & FLAIR                  & -- & 42.3 & 48.6 & 54.8 \\
nnInter. (LOGITS)     & FLAIR,SWI             & -- & 38.0 & 45.8 & 55.9 \\
nnInter. (LOGITS)     & FLAIR,T1,T2          & -- & 39.1 & 46.6 & 53.7 \\
nnInter. (LOGITS)     & FLAIR,T1,T2,SWI                      & -- & 37.4 & 45.8 & 54.7 \\
\bottomrule

\end{tabular}
\end{table*}

\begin{table*}[!t]
\centering
\caption{Percentage of detected SWI components across the entire TBI dataset for different click strategies (Overall Dice score (\%) in bracket). All experiments used T1, T2 and FLAIR. SWI was included in experiments involving the additional channel or the pretrained modality-agnostic channel. Best results highlighted in \textbf{bold} for each click strategy. (Extra Ch.: Extra channel initiated, Agn.: Agnostic channel initiated)}
\label{tab:TBI_unused modality}
\footnotesize
\begin{tabular}{@{}lccccc@{}}
\toprule
Method & Automated & BBox & BBox+1 & BBox+6 & BBox+12 \\
\midrule
\multicolumn{5}{@{}l}{\emph{Standard 1 BBox+6 clicks (largest error)}} \\
\midrule
BrainIAC (ours) -- Online
& 35.4(43.6) & 36.3(49.2) & 37.5(52.2) & 38.6(58.7) & - \\
BrainIAC (ours) + Extra Ch. -- Online
& 35.9(44.1) & 36.5(49.2) & 37.3(52.5) & 38.8(58.7) & - \\
BrainIAC (ours) + Agn. -- Online
& \textbf{47.9}(46.3) & \textbf{48.0}(50.4) & \textbf{48.4}(53.5) & \textbf{49.8}(61.0) & - \\
\midrule
\multicolumn{5}{@{}l}{\emph{1 BBox+12 alternate clicks (largest error then largest unpredicted SWI only lesion)}} \\
\midrule
BrainIAC (ours) -- Online
& 46.5(45.8) & 47.3(50.4) & 47.6(52.5) & 50.5(56.6) & 53.1(59.2) \\
BrainIAC (ours) + Extra Ch. -- Online
& 47.5(45.7) & 48.7(51.2) & 49.0(53.3) & 51.6(57.0) & 54.1(59.7) \\
BrainIAC (ours) + Agn. -- Online
& \textbf{63.6}(47.5) & \textbf{64.1}(51.3) & \textbf{64.4}(53.7) & \textbf{66.7}(58.3) & \textbf{68.9}(61.2) \\
\bottomrule
\end{tabular}
\end{table*}

\subsection{Evaluation on Other Pathologies}
\label{sec:results_extended}

To further evaluate the effectiveness of online adaptation, we also test on VES-SEG and ISLES. For these experiments, we use the \textbf{5-DB} model. These datasets are less suitable for direct comparison with foundation-model baselines such as nnInteractive and MedSAM2, since these models were trained on related databases. Therefore, in this section, we focus primarily on the improvement brought by interactive and online adaptation strategies.

\subsubsection{VES-SEG - Vestibular Schwannoma}
\label{sec:results_schwannoma}
\begin{figure}
\centering
\includegraphics[width=0.8\linewidth]{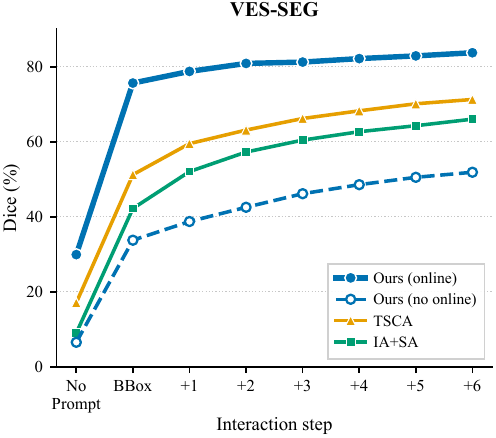}
\caption{Results on VES-SEG T1C (Dice \%) along the interaction protocol: automatic prediction (No Prompt), bounding-box initialisation (BBox), and one to six correction clicks (BBox$+1$ to BBox$+6$). Online adaptation lifts the entire trajectory far above the no-adaptation variant and the TSCA and IA+SA online-adaptation baselines.}
\label{fig:vesseg}
\end{figure}

Results on VES-SEG T1C are shown in Fig.~\ref{fig:vesseg}. On this dataset, online adaptation leads to dramatic improvements. Without online adaptation, the interactive process improves the dice score from 6.5 to 51.9, and with online adaptation it reaches 83.8. This demonstrates strong anatomical out-of-domain performance, as these tumours originate from the vestibulocochlear nerve rather than within the brain tissue on which the model is trained.  

\subsubsection{ISLES - Ischemic Stroke}
\label{sec:results_isles}

\begin{figure}
\centering
\includegraphics[width=0.8\linewidth]{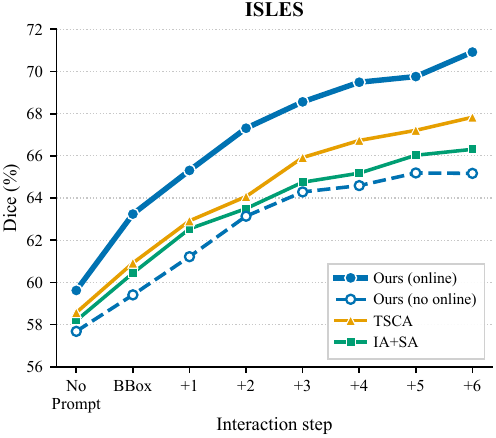}
\caption{Results on ISLES (Dice \%, FLAIR) along the interaction protocol (No Prompt, BBox, and BBox$+1$ to BBox$+6$ correction clicks). BrainIAC with online adaptation consistently surpasses the no-adaptation variant and the TSCA and IA+SA baselines across the whole trajectory.}
\label{fig:isles}
\end{figure}

The model was previously trained on ATLAS and is now evaluated on ISLES. Results can be seen in Fig.~\ref{fig:isles}. The interactive process successfully improves the result, and our online adaptation mechanism further enhances performance consistently surpassing the baseline methods.

\subsection{Qualitative Results}
\label{sec:qualitative}

Figure~\ref{fig:overview_diagram} illustrates in 3D how the predicted segmentation
evolves as the user applies an initial bounding box followed by multiple
corrective clicks. We present examples of four distinct
pathologies from different databases: TBI, MSSEG, ISLES and VES-SEG. For VES-SEG, the automated prediction produces a false positive and misses one lesion. Bounding-box prompting eliminates the false positive and enables accurate segmentation of the vestibular schwannoma, resulting in a substantial improvement in the dice score as seen in Figure \ref{fig:vesseg}. For ISLES the bounding box similarly removed false positives outside of the BBox. Overall, for each task the results demonstrate progressive refinement of the segmentation mask.

\subsection{In-Distribution Results}
\label{sec:results_id}

For completeness, Table~\ref{tab:indist} reports in-distribution results in the 5-DB setting ($C = 6$ modality channels, corresponding to 9 total input channels), without online adaptation. This is not the main focus of our study, since deployment on the same databases used for training is less challenging, and can often be better addressed by models trained for a specific task. 

BrainIAC achieves automatic segmentation performance comparable or better to database-specific nnU-Net models for 3 of the 5 databases. Interactive prompts progressively improve performance. Compared with nnInteractive, BrainIAC with BBox+6 substantially improves performance on both the WMH and ATLAS datasets (79.6 vs.\ 56.9 on WMH, and 75.1 vs.\ 71.5 on ATLAS). For ATLAS the single BBox improves the Dice from 58.2 to 68.0. We didn't apply online adaptation to BrainIAC because these experiments are in-distribution. In addition, nnInteractive was trained on the same versions of WMH and ATLAS as we use. We do not report a comparison on BraTS, as we use a different version of the dataset.

\begin{table}[!t]
\centering
\caption{In-distribution results (5-DB training, $C{=}6$ modality channels, no online adaptation). Dice (\%). Best results for each database in \textbf{bold}.}
\label{tab:indist}
\scriptsize
\setlength{\tabcolsep}{4pt}
\begin{tabular}{@{}llccccc@{}}
\toprule
Method & Interact. & MSSEG & TBI & WMH & ATLAS & BraTS \\
\midrule
\multirow{4}{*}{Ours}
 & No Prompt & 74.4 & 59.9 & 78.0 & 58.2 & 92.5 \\
 & BBox      & 74.3 & 63.6 & 78.1 & 68.0 & 92.6 \\
 & BBox+1    & 75.0 & 64.5 & 78.4 & 70.5 & 92.7 \\
 & BBox+6    & \textbf{77.7} & \textbf{66.9} & 79.6 & \textbf{75.1} & \textbf{93.2} \\
\midrule
nnU-Net  & No Prompt & 74.2 & 59.0 & \textbf{81.0} & 60.8 & 92.4 \\
nnInter. & BBox+6    & -- & -- & 56.9 & 71.5 & -- \\
\bottomrule
\end{tabular}
\end{table}

\subsection{Online Adaptation Ablation Study}
\label{sec:ablation}

Here, we conduct an ablation study to assess the contributions of the MI and PI adaptation components and evaluate their impact on the overall performance of the proposed framework. As shown in Table~\ref{tab:ablation}, both adaptation processes are important, and the best performance is obtained when they are utilised together. We also have a comprehensive ablation study in our previous 2D interactive work \citep{xu2026you}. 

\begin{table}[!t]
\centering
\caption{Ablation study of online adaptation components on TBI (FLAIR-visible lesions). MI = Mid-Interaction adaptation, PI = Post-Interaction adaptation. Dice score (\%). Best result in \textbf{bold}.}
\label{tab:ablation}
\footnotesize
\begin{tabular}{@{}cccccc@{}}
\toprule
MI & PI & No Prompt & BBox & BBox+1 & BBox+6 \\
\midrule
\xmark & \xmark & 37.9 & 39.5 & 41.1 & 47.9 \\
\cmark & \xmark & 43.9 & 49.1 & 53.2 & 61.6 \\
\xmark & \cmark & 42.2 & 46.8 & 50.9 & 59.1 \\
\cmark & \cmark & \textbf{44.3} & \textbf{50.5} & \textbf{54.3} & \textbf{62.2} \\
\bottomrule
\end{tabular}
\end{table}

\subsection{Efficiency Analysis}
\label{sec:computational}

In this section, we assess the efficiency of online adaptation by examining both its computational overhead and interaction requirements. We quantify the additional runtime introduced by the adaptation process and evaluate the amount of user input required to attain an acceptable Dice score.

\subsubsection{Per-click Latency}
\label{sec:timing_schwannoma}

We measure runtime on the VES-SEG dataset (242 samples). After each click, the mid-interaction adaptation takes about \textbf{1.19\,s}; after each sample, the post-interaction adaptation takes about \textbf{3.54\,s}. The per-click cost is small relative to the time a clinician spends placing the click itself, and the once-per-sample update adds only a few seconds, hence, the additional latency introduced by online adaptation is acceptable and the method remains efficient for interactive clinical use.

Furthermore, when efficiency is critical, clinicians can rely on post-interaction adaptation alone, which still significantly improves the result without adding any latency to the interactive process itself.

\subsubsection{Click Efficiency}
\label{sec:click_efficiency}

To quantify how much user effort is saved by online adaptation, we measure the number of correction clicks required to reach a target Dice on each case (up to a maximum of 15 clicks per case); cases that never reach the threshold are counted as 15 clicks. We evaluate on four databases and set the target Dice according to the difficulty of each pathology: 0.70 on MSSEG, 0.80 on VES-SEG, and 0.65 on the more challenging ISLES and TBI datasets. Alongside the average number of clicks, we also report the fraction of cases that actually reach the threshold within the 15-click budget. Table~\ref{tab:click_efficiency} summarizes the results.

Online adaptation reduces the number of clicks required on every database while simultaneously enabling more cases to reach the target Dice. The effect is largest on VES-SEG, where adaptation cuts the average from 12.81 to 2.67 clicks and raises the fraction of cases reaching 0.80 Dice from 31.8\% to 90.1\%. It is also substantial on TBI (10.35$\rightarrow$6.78 clicks; 39.9\%$\rightarrow$69.8\%), MSSEG (6.66$\rightarrow$3.81 clicks; 71.7\%$\rightarrow$96.2\%), and ISLES (5.46$\rightarrow$3.43 clicks; 64.3\%$\rightarrow$85.7\%). Across all four databases, adaptation therefore delivers acceptable segmentations with fewer interactions on a substantially greater proportion of cases. 
\begin{table}[!t]
\centering
\caption{Click efficiency at fixed Dice thresholds. For each database we report the fraction of cases that reach the target Dice within a 15-click budget and the average number of correction clicks per case (cases that never reach the threshold are counted as 15 clicks). Thresholds: MSSEG = 0.70, VES-SEG T1C = 0.80, ISLES = 0.65, TBI FLAIR = 0.65. Best result in each row is shown in \textbf{bold}.}
\label{tab:click_efficiency}
\footnotesize
\begin{tabular}{@{}lcccc@{}}
\toprule
& \multicolumn{2}{c}{Avg clicks} & \multicolumn{2}{c}{Reached threshold (\%)} \\
\cmidrule(lr){2-3} \cmidrule(l){4-5}
Dataset & w/ Online & w/o Online & w/ Online & w/o Online \\
\midrule
MSSEG   & \textbf{3.81} & 6.66 & \textbf{96.2} & 71.7 \\
VES-SEG & \textbf{2.67} & 12.81 & \textbf{90.1} & 31.8 \\
ISLES   & \textbf{3.43} & 5.46 & \textbf{85.7} & 64.3 \\
TBI     & \textbf{6.78} & 10.35 & \textbf{69.8} & 39.9 \\
\bottomrule
\end{tabular}
\end{table}

\noindent\textbf{Click efficiency on recovering undetected lesion components:}
As discussed in section \ref{sec:MSSEG}, MSSEG subjects typically contain a large number of small, scattered lesions, hence, a common failure mode is that entire lesion components are missed. To probe this behaviour directly and to test how efficiently clicks can recover these components, we design a component-oriented interaction protocol that differs from the protocol used above. For each case, no bounding box is provided, and every correction click is placed (at a random location) inside a ground-truth component that the current prediction has failed to detect. If at any step no undetected component remains to be clicked, the case is stopped early. We consider two variants: a \emph{10-click budget}, which stops each case after at most 10 clicks, and an \emph{unlimited} setting, which keeps clicking until no undetected component remains. These experiments isolate the model's ability to recover missed components and let us track how many components remain undetected as clicks accumulate. We run this protocol on the MSSEG dataset (53 subjects), comparing the model with and without online adaptation.

Table~\ref{tab:undetected} reports the results. For the limited click setting, with online adaptation, far fewer lesion components remain undetected both before and after the interaction steps. The `before' count refers only to the current sample; because the online-adapted model also retains information from previously seen samples, it starts each case with fewer undetected components. Given enough clicks, every ground-truth component can eventually be recovered, so under the unlimited setting, both models reach zero undetected components. The difference is the effort required: without online adaptation, the model does not improve, so almost every missed component must be clicked individually, needing 20.94 clicks per case on average. With online adaptation, only 3.15 clicks per case are needed --- even fewer than under the 10-click budget (4.00). This happens because removing the max click budget allows the model to receive all the corrective clicks it requires from the early cases, and strong early adaptation carries over to later images, which then start with far fewer undetected components and require almost no clicks. In summary, the improvement obtained on the first several images has a strong positive influence on all subsequent images, so the clinician does not need to click on every missed component: online adaptation trains the model to find the remaining ones on its own.

\begin{table}[!t]
\centering
\caption{Undetected lesion-component analysis on the held-out MSSEG dataset (53 subjects). Each correction click is placed inside a currently undetected ground-truth component. We report the average number of undetected lesion components pre and post-clicks, together with the average clicks used under both a  10-click budget and under an unlimited-click setting. With unlimited clicks every component is eventually recovered (zero undetected) for both models, but online adaptation reaches this with far fewer clicks.}
\label{tab:undetected}
\footnotesize
\begin{tabular}{@{}lcc@{}}
\toprule
Metric & w/ Online & w/o Online \\
\midrule

\multicolumn{3}{@{}l}{\textit{10-click budget (max 10 clicks/case)}} \\
\quad Avg. undetected -- pre-clicks       & \textbf{8.26}  & 23.75 \\
\quad Avg. undetected -- post-clicks   & \textbf{2.51}  & 13.87 \\
\quad Avg. clicks used               & \textbf{4.00}  & 7.96 \\
\midrule
\multicolumn{3}{@{}l}{\textit{Unlimited clicks (until all detected)}} \\
\quad Avg. undetected -- pre-clicks       & \textbf{5.75}  & 23.75 \\
\quad Avg. undetected -- post-clicks   & 0  & 0 \\
\quad Avg. clicks used               & \textbf{3.15}  & 20.94 \\
\bottomrule
\end{tabular}
\end{table}

\subsubsection{Data Efficiency}
\label{sec:adaptation_curve}

\begin{figure}
\centering
\includegraphics[width=0.95\linewidth]{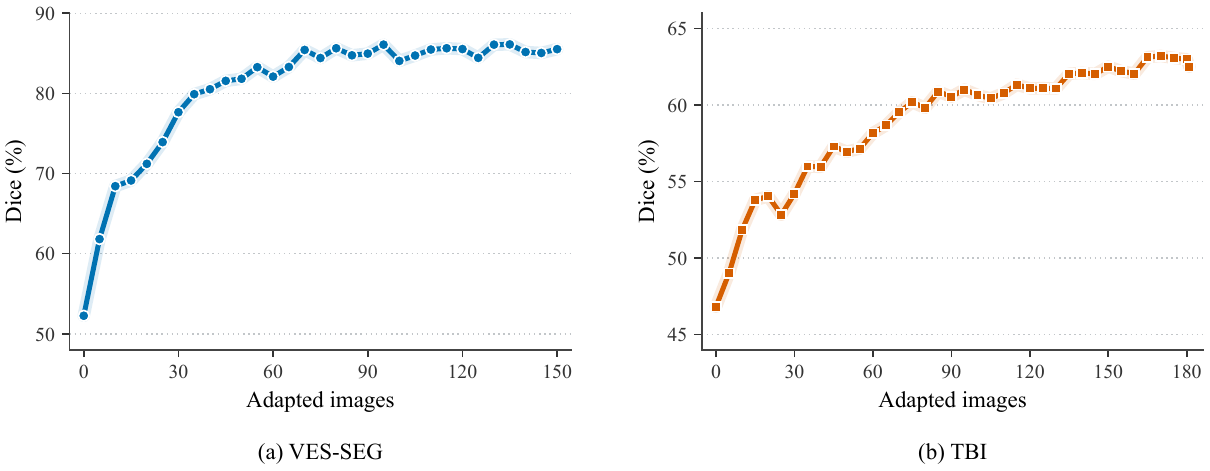}
\caption{Online-adaptation trajectory on (a) VES-SEG and (b) TBI. The $x$-axis is the number of adaptation images processed so far, and the $y$-axis is the average Dice on the held-out evaluation split}
\label{fig:adaptation_curve}
\end{figure}

To analyse how efficiently our online adaptation improves on an unseen pathology, we follow a held-out protocol on VES-SEG and TBI. For each database the data is split into an adaptation split and a held-out evaluation split, and online adaptation is performed only on the adaptation split. The VES-SEG splits contain 150 adaptation and 92 held-out images, and the TBI splits contain 181 adaptation and 100 held-out images. The model is adapted sequentially on the adaptation split (one image at a time), and after every five adaptation images we evaluate it on the entire held-out split, providing a bounding box and six correction clicks (BBox+6) per held-out image without adapting to them. Figure~\ref{fig:adaptation_curve} plots the average Dice for BBox+6 interaction, over the held-out set against the number of adaptation images processed so far.

On both databases the held-out Dice improves continually as more adaptation images are processed and then converges to a stable level. On VES-SEG, the Dice climbs steeply from about 52\% to 77\% after around 30 adaptation images, reflecting how quickly adaptation recalibrates the model to this new distribution. On TBI, which is harder and shifts more gradually, the improvement is slower but steady: the Dice keeps rising from about 47\% to roughly 60\% after around 70 adaptation images. In both cases, the model continues to improve well beyond the first few images, confirming that the benefit of online adaptation accumulates with continued deployment rather than saturating after a small number of images.

\section{Discussion}
\label{sec:discussion}

In this work we developed a single 3D framework for interactive brain lesion segmentation that unifies three capabilities: (i) Handling heterogeneous modality sets, including new modalities at inference; (ii) 3D interactive segmentation with bounding-box and click prompts that preserves fully automatic prediction when no prompt is given; and (iii) an online adaptation mechanism combining Mid-Interaction adaptation and Post-Interaction adaptation.
We evaluated it on seven brain MRI databases that span multiple pathologies and modality configurations. We proceeded to analyse a variety of metrics including the final accuracy, interaction effort, data efficiency of adaptation, and the ability to recover missing lesion components with a focus on out of distribution testing. In this section we discuss these results.

\textbf{Interactive prompting} is effective on every database. A single bounding box already significantly assists in predicting a focal (often undetected) lesion as exemplified on VES-SEG, which represents a novel pathology not encountered by the model during training, the interactive process improves Dice from 6.5 to 33.8 even prior to adaptation and subsequent correction clicks add further gains.

To better reflect real-world clinical practice, we also evaluated other interaction strategies for different types of lesion. For MSSEG, these included multiple bounding boxes for scattered lesions. For TBI, we evaluated selecting components seen on different modalities, SWI and FLAIR. These experiments demonstrate benefits in different aspects of segmentation performance, including overall Dice and the detection of SWI-visible lesion components.Unlike foundation models such as MedSAM2 and nnInteractive, our model retains a fully automatic mode (No Prompt); combined with online adaptation, this means that after several images some cases already satisfy the clinician without a single click. 

\textbf{Online adaptation has a strong influence on performance.} This is one of our main contributions. Adaptation improves the BBox+6 Dice on every held-out database,  $+4.3$ on MSSEG FLAIR ($68.5\rightarrow72.8$), $+14.3$ on TBI FLAIR ($47.9\rightarrow62.2$), $+5.7$ on ISLES FLAIR ($65.2\rightarrow$70.9) , and $+31.9$ on VES-SEG ($51.9\rightarrow83.8$). To our knowledge, this is one of the first online adaptation methods for \emph{3D} interactive segmentation; prior online adaptation for interactive segmentation, including our own, was restricted to 2D~\citep{xu2026you}, and existing medical 3D interactive methods keep the model frozen at test time \citep{isensee2025nninteractive, ma2025medsam2}.

 User prompts combined with online adaptation substantially improve performance through two mechanisms. First, corrected predictions are used as pseudo-labels to update the model immediately, reducing the likelihood of repeating similar errors on subsequent cases. Second, the update absorbs the correction into the weights, as a result, the  spatially local click can produce a relative global influence. Consequently, online adaptation improves both segmentation accuracy and annotation efficiency for current and future images, as demonstrated by the results.

\textbf{Efficiency} is what makes this improvement clinically usable. Online adaptation adds extra time to the interactive process. In interactive segmentation, the true cost is clinician time, dominated by the number of prompts needed and the latency of each response. Online adaptation reduces the number of prompts needed to reach a relatively good performance (for example 2.67 vs.\ 12.81 clicks on VES-SEG and 3.81 vs.\ 6.66 on MSSEG). The natural concern is added latency, but our updates are light, 1.2\,s for the per-click mid-interaction update and a few seconds for the once-per-case post-interaction update, which are negligible beside the time a clinician spends placing a prompt and far cheaper than heavier adaptation schemes such as IA+SA~\citep{kontogianni2020continuous}.

\textbf{Handling multiple modalities} is likewise essential for brain MRI, where the available sequences vary across sites and protocols, and our results show that they should be fused \emph{inside} the model rather than after it. The Multi-Unet backbone ingests all available modalities jointly and can benefit from extra channels, especially at the start of each case before any prompt (adding T1 to FLAIR on MSSEG improves the No-Prompt Dice from 64.7 to 66.1). In contrast, the simpler alternative of keeping a single-modality model and combining its per-modality predictions post hoc proves unreliable: with nnInteractive, the LOGITS strategy degrades from 54.8 (FLAIR) to 53.7 (FLAIR, T1, T2) at BBox+6. The size of the multi-modal gain depends on how much genuinely complementary information a modality carries, and sometimes one sequence is already near-sufficient, but joint modeling gives the clinician the option to utilise all available diagnostic information, something previous interactive methods, such as nnInteractive and MedSAM2, cannot do.

A static model, including our previous Multi-UNet work~\citep{xu2024feasibility}, cannot easily exploit information from an unseen modality. By initiating the modality-agnostic channel~\citep{addison2025modality}, BrainIAC can incorporate modalities that were never observed during training. This is exemplified in TBI, where BrainIAC achieves an improved dice score alongside a substantial improvement in detecting SWI-only lesion components under online adaptation when incorporating SWI. In contrast, simply adding a new channel with randomly initialized weights does not provide meaningful improvement. The large representation gap introduced by random initialization prevents the model from effectively utilizing the unseen modality, even with online adaptation and interactive guidance. 

These results highlight the importance of a properly initialized modality-agnostic representation for incorporating unseen modalities alongside modalities seen during training at inference. These three mechanisms point towards a single robust model that can be placed in an unfamiliar clinical environment, with an unfamiliar modality set, and improve itself with with online learning via user interaction..

\textbf{Study limitations:} Our study has several limitations. Processing full $192\times192\times192$ volumes is well suited to segmentation of brain lesions, as brain volumes are of a similar size; applying the method to other anatomies may require methods such as sliding window inference. The large $192$ volume also affects training efficiency. Our evaluation uses simulated clicks and boxes; although the simulation follows common protocols, a user study with radiologists would better capture the variability of real interaction. Finally, the proposed framework is designed for binary lesion segmentation. Extending it to multi-class tasks, such as delineating tumour sub-regions or classifying distinct connected tumour components arising from different pathological processes, longitudinal tracking, and additional prompt types such as scribbles, are all natural next steps.

\section{Conclusion}

\label{sec:conclusion}
We presented BrainIAC, a unified framework for 3D interactive brain lesion segmentation that can handle heterogeneous MRI modalities, support different interaction strategies (bounding boxes, clicks, and fully automatic prediction), and perform online adaptation that learns from user interactions to continuously adapt to new data distributions. A single model can thus be deployed across diverse clinical settings with varying modalities including imaging modalities not encountered during training and a range of pathologies, including previously unseen pathologies.Experiments on seven brain MRI databases demonstrate that the three components produce synergistic benefits, offering insights for building 3D segmentation models in medical imaging. Future work will explore parameter-efficient adaptation, additional prompt types such as scribbles, and other online adaptation methods.

\printcredits

\section*{Acknowledgements}

APA is supported by the Oxford-Bellhouse Graduate Scholarship, jointly funded by Magdalen College and the University of Oxford.  ZL is supported by scholarship provided by the EPSRC Doctoral Training Partnerships programme [EP/W524311/1]. HA is supported by a scholarship via the EPSRC Doctoral Training Partnerships programme [EP/W524311/1, EP/T517811/1].

\bibliographystyle{cas-model2-names}
\bibliography{references}

\end{document}